\documentclass[10pt,twocolumn,letterpaper]{article}

\usepackage{iccv}
\usepackage{times}
\usepackage{epsfig}
\usepackage{graphicx}
\usepackage{amsmath}
\usepackage{amssymb}

\usepackage[pagebackref=true,breaklinks=true,letterpaper=true,colorlinks,bookmarks=false]{hyperref}

\iccvfinalcopy 

\def\iccvPaperID{8} 

\ificcvfinal\fi

\begin{document}

\title{From LAION-5B to LAION-EO:\\Filtering Billions of Images Using Anchor Datasets\\for Satellite Image Extraction}

\author{Mikolaj Czerkawski, Alistair Francis\\
$\Phi$$$-lab\\
European Space Agency\\
{\tt\small mikolaj.czerkawski@esa.int}
}

\maketitle
\ificcvfinal\thispagestyle{empty}\fi


\begin{abstract}
   Large datasets, such as LAION-5B, contain a diverse distribution of images shared online. However, extraction of domain-specific subsets of large image corpora is challenging. The extraction approach based on an anchor dataset, combined with further filtering, is proposed here and demonstrated for the domain of satellite imagery. This results in the release of LAION-EO, a dataset sourced from the web containing pairs of text and satellite images in high (pixel-wise) resolution. The paper outlines the acquisition procedure as well as some of the features of the dataset. Access at \url{https://huggingface.co/datasets/mikonvergence/LAION-EO}.
\end{abstract}


\section{Introduction}

    Large collections of images have been instrumental for building the recent generative models, such as DALLE~\cite{Ramesh2021}, Imagen~\cite{Saharia2022}, or StableDiffusion~\cite{Rombach_2022_CVPR}. LAION-5B~\cite{Schuhmann2022laionb} is one of the largest open-source image datasets containing pairs of image and text and was used to train some of the state-of-the-art generative models, including StableDiffusion~\cite{Rombach_2022_CVPR}.

    However, the datasets used for training the general text-to-image generative models capture a correspondingly wide domain of visual data. Hence, they may be less than ideal for individual domains, such as those involving processing of satellite-images. There are several potential benefits of more specialised vision-language datasets derived from general corpora. First, they can provide a way to fine-tune existing models on the domain of interest in order to improve the prediction quality further, tangentially explored in the works on subject-driven image editing~\cite{hu2022lora,gal2023an,Ruiz_2023_CVPR}. Secondly, they offer insight into the representation of the given domain within the parent distribution of data and consequently, the representation learned by models trained on that broader distribution.

    This work demonstrates an approach to filtering a large-scale dataset of LAION-5B to extract a satellite image subset from it and analyses the resulting collection of images. Several attempts have been made to filter large datasets like LAION-5B~\cite{Radenovic_2023_CVPR,maini2023tmars}, however, the previous studies focused on improving the general quality of the dataset as opposed to extracting a specific domain like herein.

\section{LAION-5B for Satellite Images}

    A common way to navigate the space of very large datasets relies on embeddings of a general vision-language model, such as CLIP~\cite{Radford2021learning}, and this has been done for the official demonstration of LAION-5B dataset~\cite{Schuhmann2022laionb}. The CLIP model, which is trained with a contrastive loss, paves the way for assessing pair-wise similarity of image and text samples. By encoding a piece of text or an image into the embedding space of CLIP, it is possible to identify meaningful nearest neighbours by comparing to the embedding vectors of other samples. However, it has not yet been demonstrated how the entire domain of satellite images can be identified.

    \subsection{Anchor Dataset}

        \begin{figure}
            \centering
            \includegraphics[width=\columnwidth]{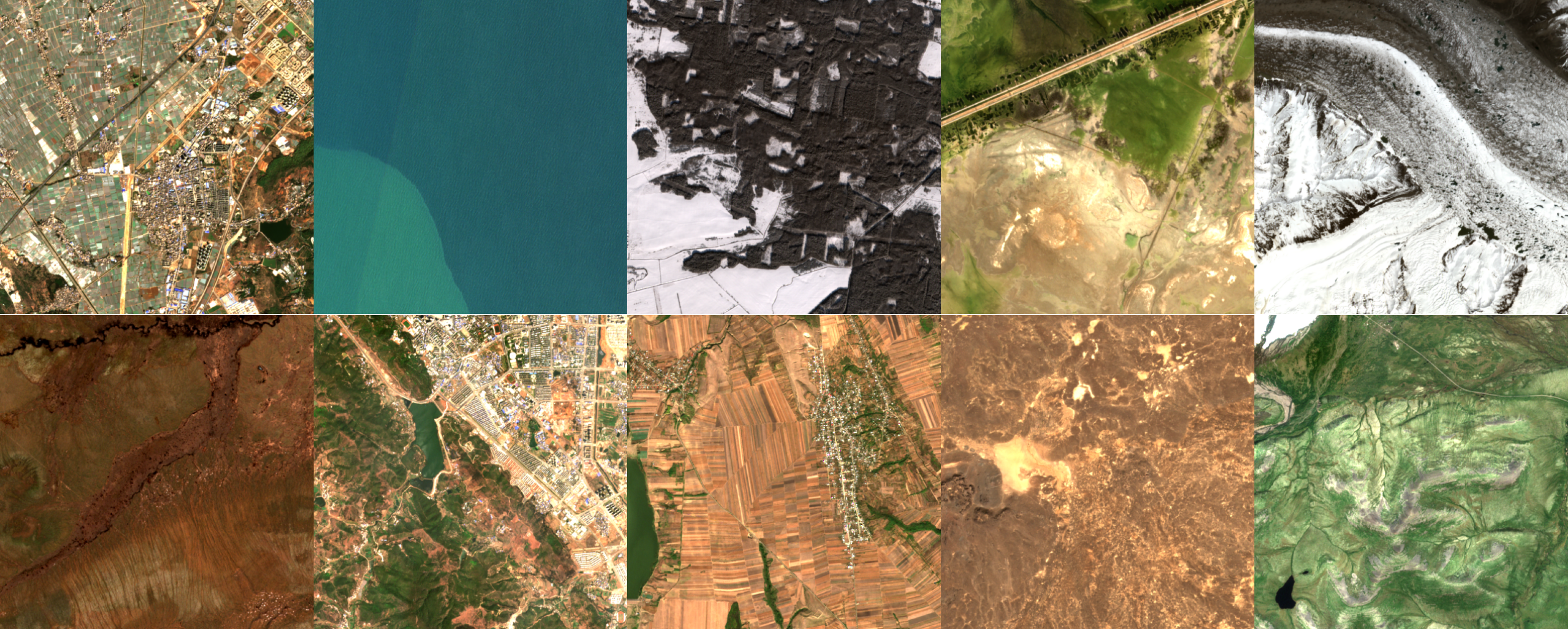}
            \caption{Random samples from the anchor dataset derived from CloudSEN12.}
            \label{fig:cloudsen12_example}
        \end{figure}
        
        In this work, the technique of using a reference anchor dataset is introduced. An anchor dataset is selected as a representative subset of the domain of interest. By iterating over each sample in the anchor dataset, a subset of LAION-5B can be identified that resembles the original dataset, and to some degree, the domain of similar-looking images in the dataset. For that reason, it is important for the anchor dataset to capture a faithful representation of the domain of interest, the satellite images.
    
        In this case, Sentinel-2 images in the resolution of 512 pixels by 512 pixels and minimal (mostly none) cloud coverage are used. To enable that, extracted are the cloud-free samples from the training subset of the CloudSEN12 dataset~\cite{Aybar2022} (the samples with high quality and scribble annotations are organised into training, validation, and test sets in the official release). The selection is motivated by two key features of the dataset, despite being designed for the task of cloud detection: i) it contains Sentinel-2 images with large patches of 512 pixels, larger than other similar datasets with Sentinel-2 data (WorldStrat contains 160 by 160 pixels~\cite{cornebise2022open}, BigEarthNet contains 120 by 120 pixels~\cite{Sumbul2019}) and ii) the data is sampled globally, with 9,880 unique locations. Furthermore, the use of an official training subset makes it possible to later evaluate on the test subset for the same dataset, for example, when LAION-EO dataset is used as a substitute source of training samples.

        The data is obtained in the standard Sentinel-2 level 2A format, with values ranging between 0 and 10,000. To make the images suitable for the CLIP model (which has been trained with images from the web in a standard format), the following normalization approach is conducted. First, the images are divided by 10,000 to fit into the standard range of 0-1. Then, if the image is too dark (which is often the case for images with no snow or clouds), a gain is applied. Specifically, if the mean reflectance is less than 0.25, the entire image is multiplied by the ratio of 0.5 to the image mean (clipped for maximum gain of 8). This approach may be unconventional in the Earth Observation pipelines, but should be taken in order to increase the similarity between the CloudSEN12 anchor images and the images that may potentially be found in LAION-5B. Examples of samples from the used CloudSEN12 subset and the effect of normalization are shown in Figure~\ref{fig:cloudsen12_example}.

        \begin{figure}
            \centering
            \includegraphics[width=\columnwidth]{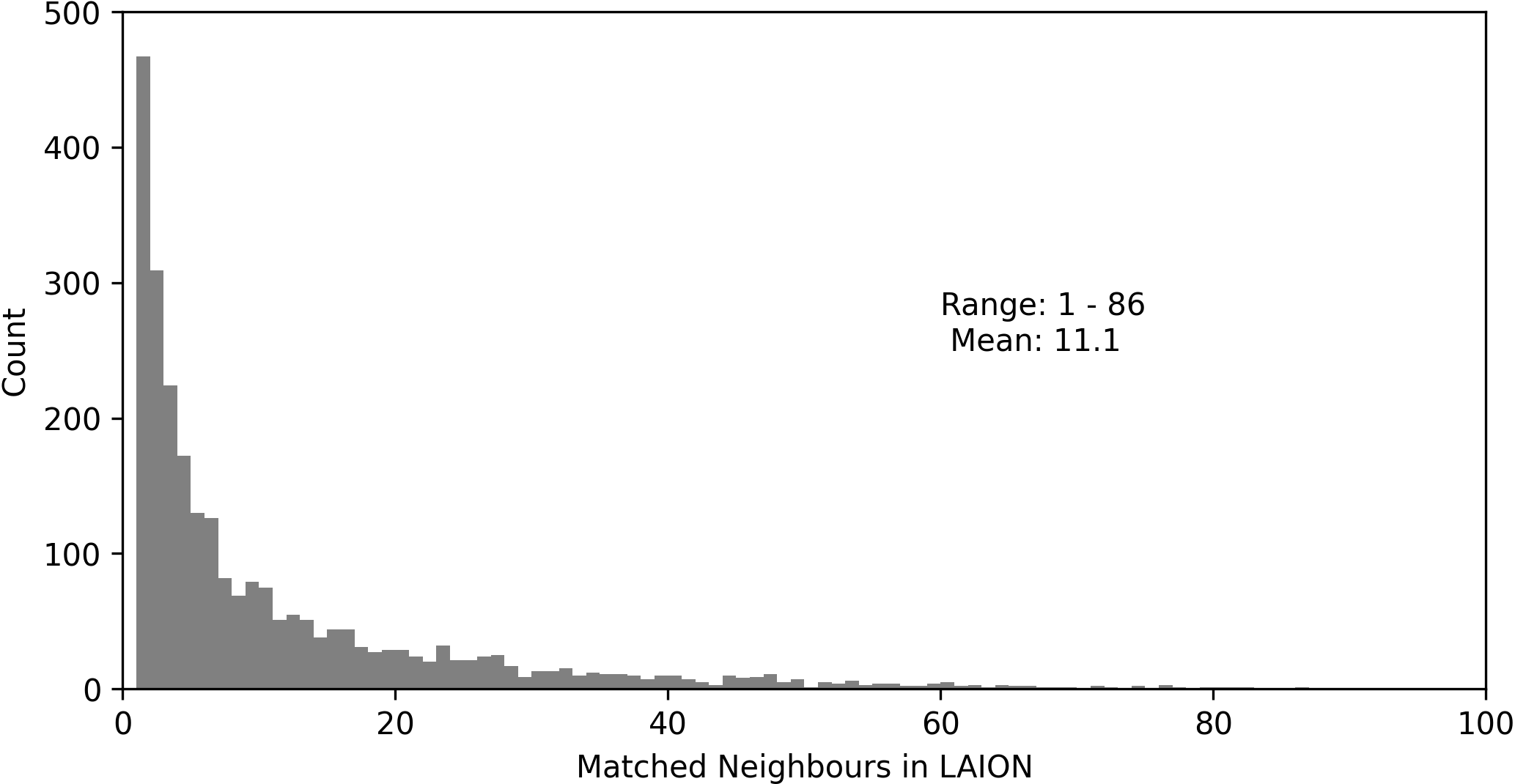}
            \caption{Distribution of the number of neighbours found for a single anchor image.}
            \label{fig:match_count}
        \end{figure}

    \subsection{Nearest Neighbour Search}

        As described in the appendices C.2 and C.3 of the LAION-5B paper~\cite{Schuhmann2022laionb}, the dataset images can be compressed into a large k-NN index of CLIP embeddings. The nearest neighbour search has been implemented based on FAISS search algorithm (using approximate similarity)~\cite{Johnson2021}. The $\texttt{clip-retrieval}$ open-source tool~\cite{beaumont2022clipretrieval} is employed to find 100 nearest neighbours with FAISS in the LAION-5B dataset for each of the 3,456 anchor images extracted from CloudSEN12. Notably, each of the query result may contain a certain number of duplicates, so 100 corresponds to the maximum number of neighbours found.

        The process involved querying 100 nearest neighbour images, ignoring duplicates, checking if it is possible to download the image, and if so, checking if the image has at least 256 pixels in both width and height.

        As a result, 28,572 neighbour images have been extracted from LAION-5B, derived from 2,582 out of the initial 3,456 anchor images (74.71\%). This means that for 874 images, the results contained images that were too small or inaccessible. The average number of images extracted based on a single anchor image is 11.1, as shown in Figure~\ref{fig:match_count}.

    \subsection{Filtering}

        \begin{figure}
            \centering
            \includegraphics[width=\columnwidth]{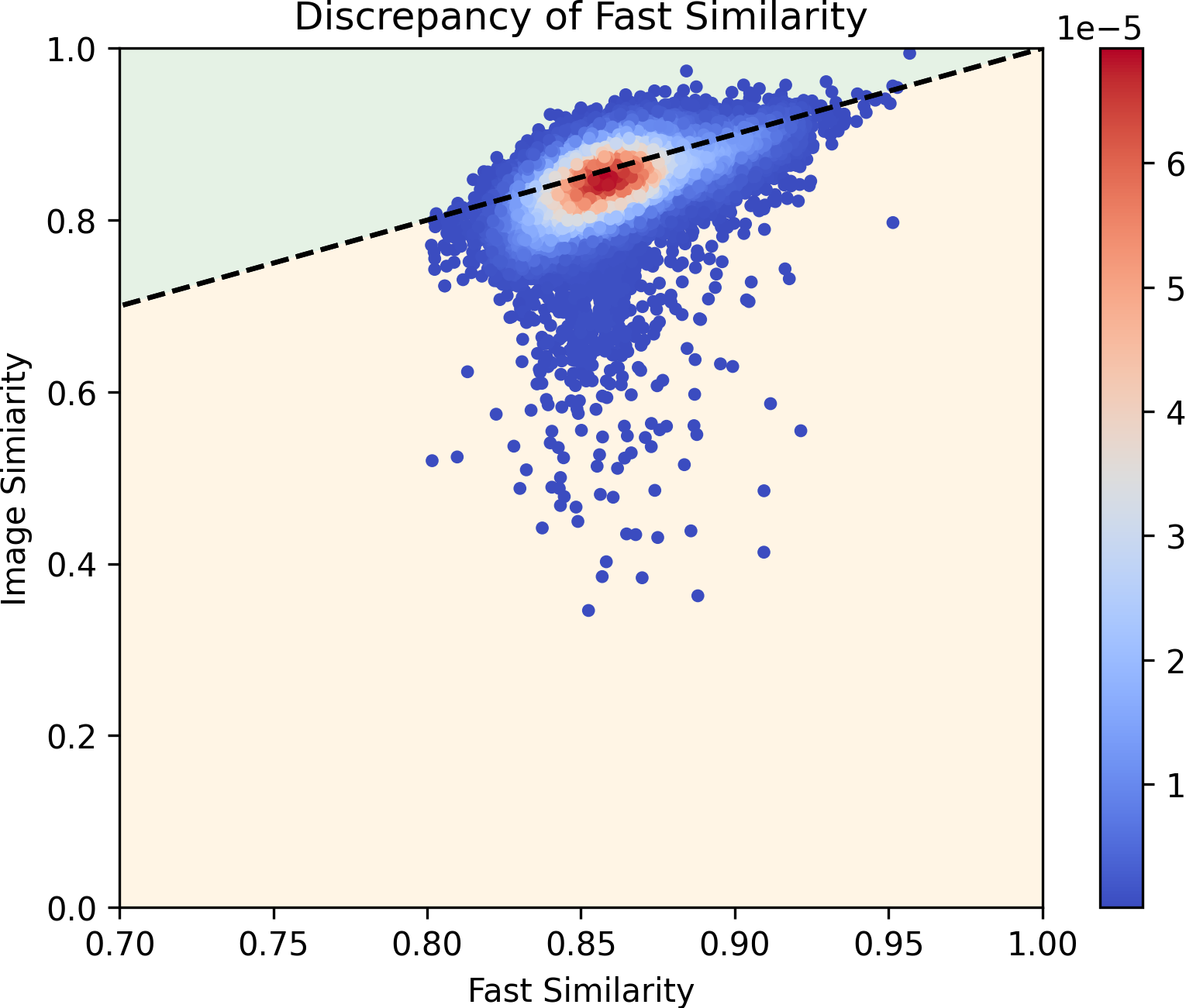}
            \caption{Comparison of the actual CLIP image similarity and the fast similarity computed by FAISS within the nearest neighbour pipeline. Colour of the markers corresponds to the fraction of all samples occupying given region.}
            \label{fig:scatter-similarity}
        \end{figure}

        \begin{figure*}
        \centering
            \includegraphics[width=\textwidth]{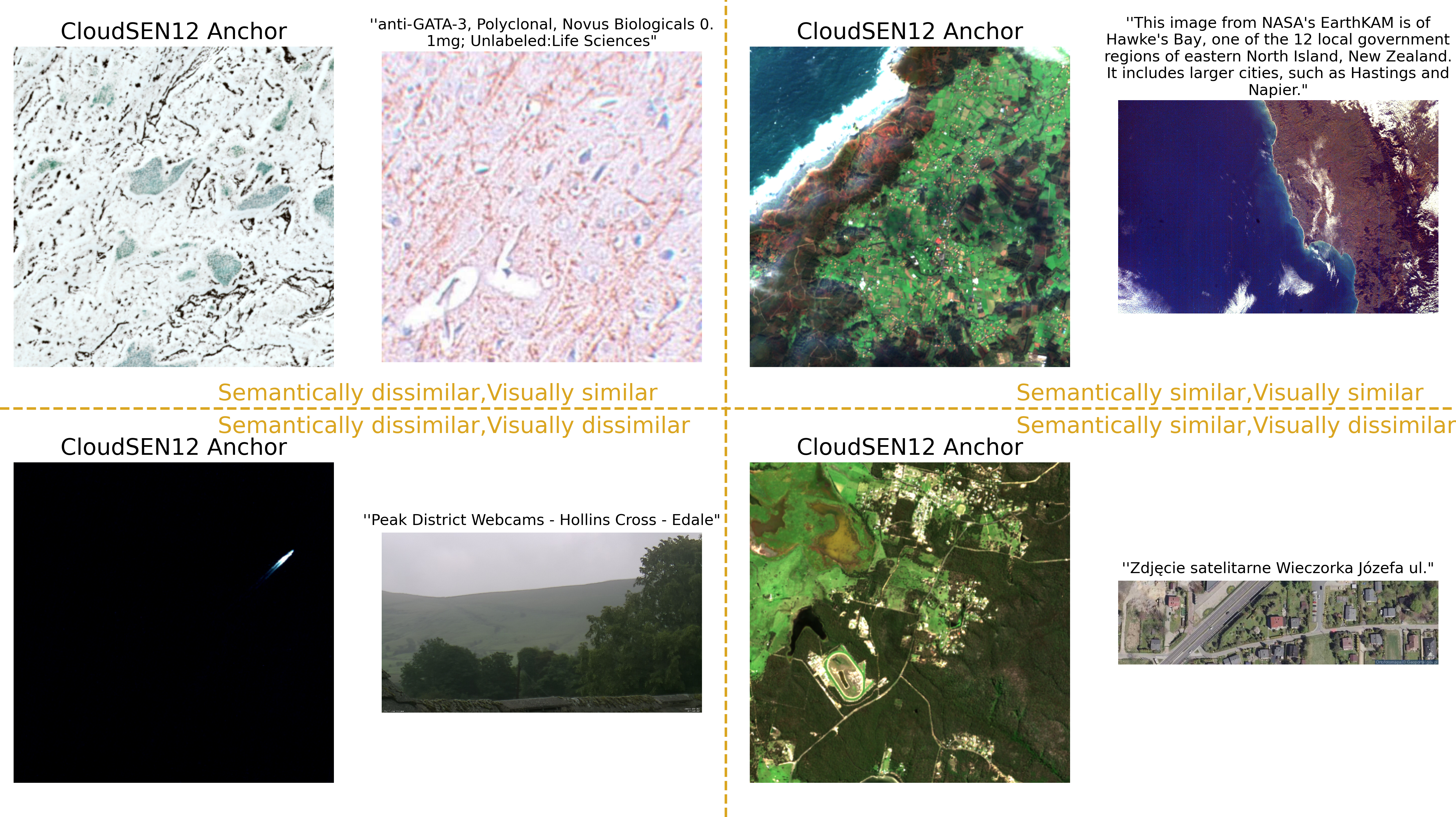}
            \caption{Types of matches found in each threshold quadrant, corresponding to the four areas in Figure~\ref{fig:thresholding}. These correspond to the four scenarios based on whether the extracted image matches the anchor image or semantic text ("a satellite image").}
            \label{fig:thresholding_quadrants}
        \end{figure*}

        The acquired 28,572 images are likely to include some incorrect samples due to several reasons. First, the embeddings are compared using FAISS~\cite{Johnson2021}, which introduces an approximation error. Second, not all images similar to satellite images are satellite images themselves. To minimize this effect, another filtering stage is carried out by recomputing the CLIP similarity of each found image to (i) the anchor image (visual similarity), and (ii) a text prompt "a satellite image" (semantic similarity), attempting to reduce the two types of errors described above.

        Figure~\ref{fig:scatter-similarity} shows the divergence between the computed fast CLIP similarity and the actual CLIP image similarity to the anchor image, unveiling a number of examples where FAISS overestimates the actual image similarity (orange area). The recomputed values of image similarity are used consequently for further filtering and the initial fast similarity is ignored.

        \begin{figure}
        \centering
            \includegraphics[width=\columnwidth]{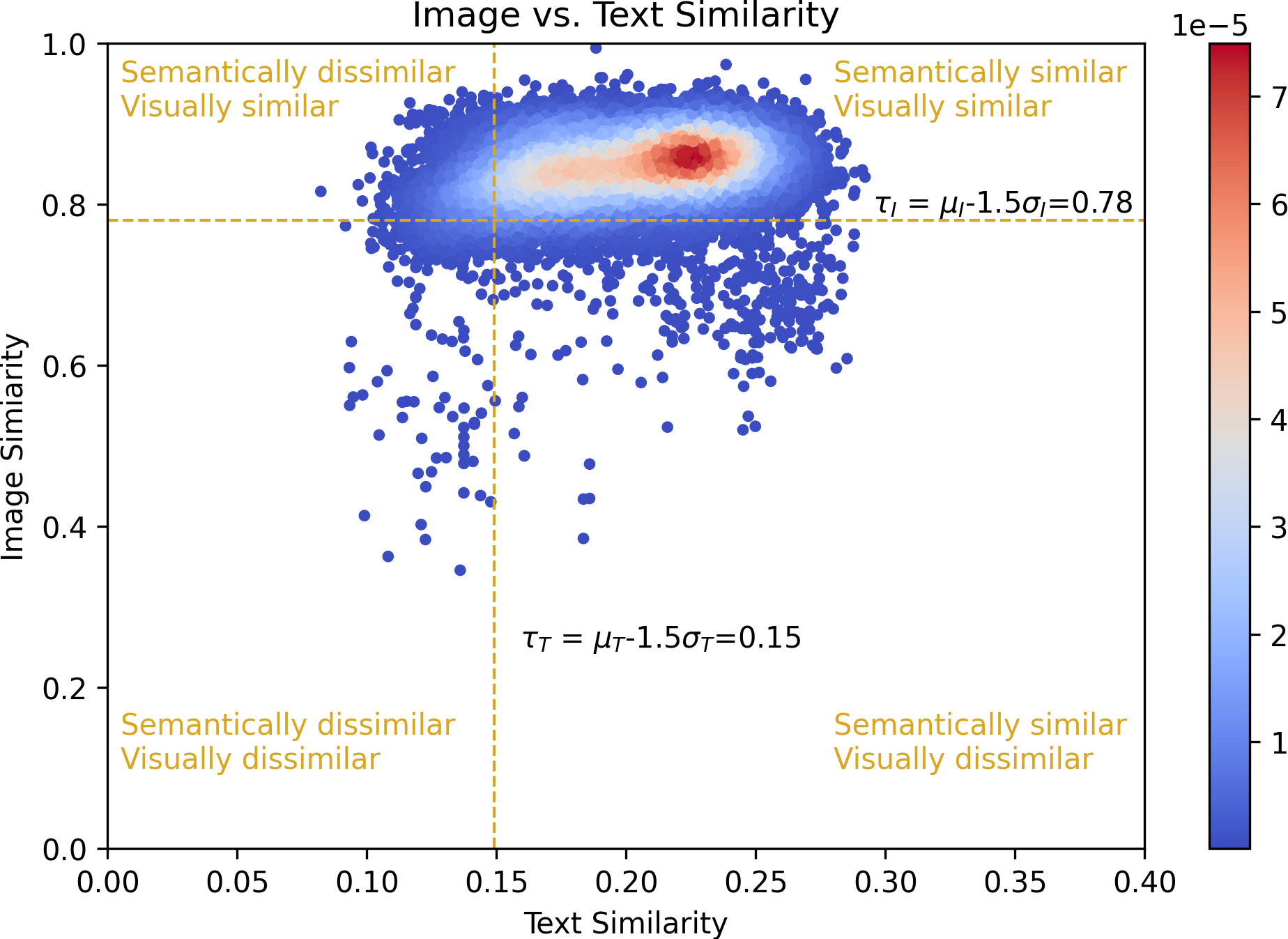}
            \caption{Distribution of image and text similarity used for filtering with corresponding threshold values based on the mean and standard deviation of each distribution.}
            \label{fig:thresholding}
        \end{figure}

        The distribution of text similarity (to the "a satellite image" text) and image similarity (to the anchor image) is shown in Figure~\ref{fig:thresholding}. As in Figure~\ref{fig:scatter-similarity}, many samples with low image similarity can be found as well as a set of samples with low text similarity. These samples with either or both similarity values too low can be filtered out using thresholds ($\tau_I$ for image similarity and $\tau_T$ for text similarity), marked by golden lines in Figure~\ref{fig:thresholding}. The exact values of thresholds have been set heuristically to one-and-a-half standard deviation below the mean of each distribution. The errors corresponding to the four quadrants separated by threshold values are shown in Figure~\ref{fig:thresholding_quadrants}, where the found image is visually similar to the anchor, but not to a satellite image (top left), similar in both respects (top right), or dissimilar visually (bottom).
    
        \begin{figure*}
            \centering
            \includegraphics[width=\textwidth]{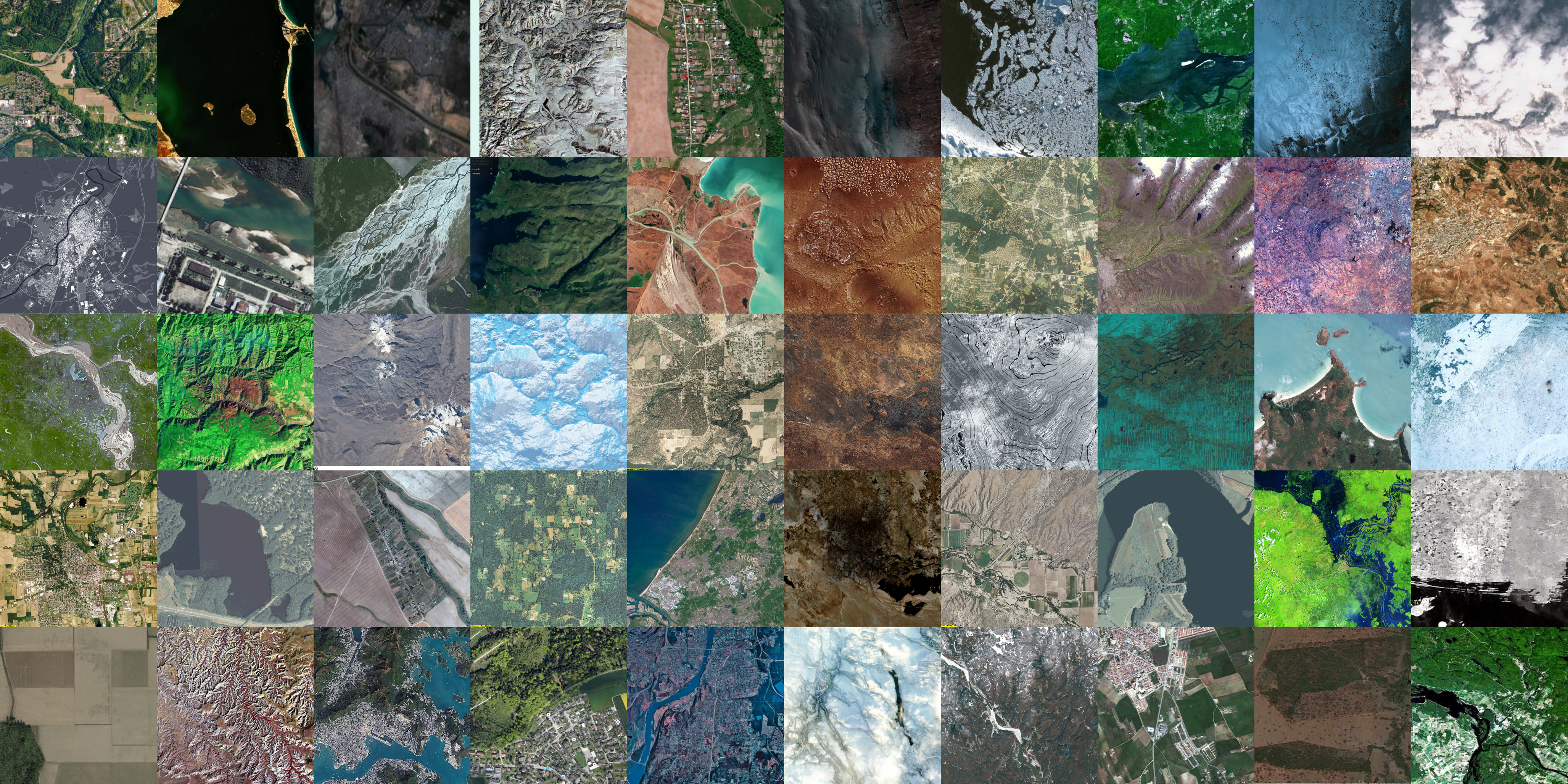}
            \caption{Random samples from LAION-EO with English captions and high semantic similarity (over 0.22), cropped to 1:1 ratio. There are 4,116 high-quality images under this criterion.}
            \label{fig:collage}
        \end{figure*}
    
        \begin{figure}
            \centering
            \includegraphics[width=\columnwidth]{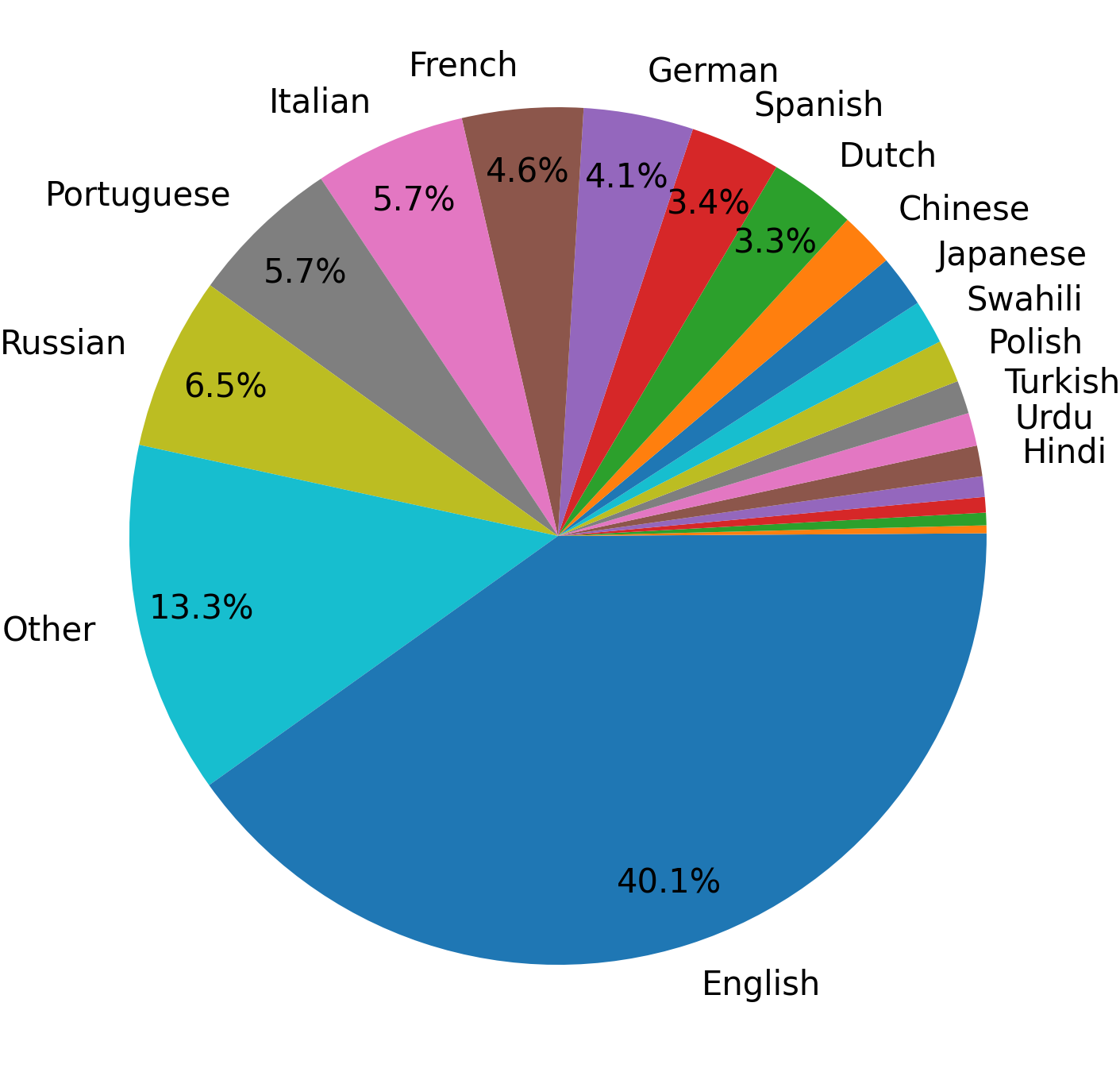}
            \caption{Distribution of language detected by the XLM-RoBERTa model~\cite{languageModel}.}
            \label{fig:lang-chart}
        \end{figure}

    \subsection{Final Product: LAION-EO}

        The final dataset is obtained after applying the image and text similarity thresholds and removing 273 duplicate images. The resulting is a set of 24,933 samples (from initial 28,572), with mean height of 633.0 pixels (up to 9,999) and mean width of 843.7 pixels (up to 19,687). Examples of square crops from random samples in the dataset are shown in Figure~\ref{fig:collage}.
    
        The dataset contains English captions (40.1\%) as well as other common languages in LAION-5B. This has been approximated by applying an open-source language detection model~\cite{languageModel}, as shown in Figure~\ref{fig:lang-chart}.
    
\section{Conclusion}

    Powerful general visual generative models for satellite images require diverse data to train on. Currently, the majority of Earth observation datasets are collected manually with a specialised acquisition pipeline. In contrast, LAION-EO is a dataset consisting of satellite images shared on web by humans, accompanied by text. It represents a fundamental change of perspective, where the images are collected because they have been found meaningful by the internet users, not necessarily experts. LAION-EO can open doors to training or fine-tuning vision-language models focused on Earth Observation data and understanding the representation of satellite images present online.

    The approach of using an anchor dataset for nearest neighbour search allows to elevate existing satellite image datasets to factorise and inspect the 5 bilion images contained in LAION-5B. This was followed by a filtering stage, where the embedding of a reference text "a satellite image" was used to clean up the dataset and remove images that do not resemble satellite images.

    Several directions to developing the LAION-EO dataset and increasing the quality of the contained data include expanding the coverage of the anchor dataset and applying more complex filtering approaches. LAION-5B likely contains even more high-quality Earth Observation images than the ones extracted in the prototype version of LAION-EO.

{\small

\begin{thebibliography}{10}\itemsep=-1pt

\bibitem{Aybar2022}
Cesar Aybar, Luis Ysuhuaylas, Jhomira Loja, Karen Gonzales, Fernando Herrera,
  Lesly Bautista, Roy Yali, Angie Flores, Lissette Diaz, Nicole Cuenca, Wendy
  Espinoza, Fernando Prudencio, Valeria Llactayo, David Montero, Martin
  Sudmanns, Dirk Tiede, Gonzalo Mateo-Garc{\'i}a, and Luis G{\'o}mez-Chova.
\newblock Cloudsen12, a global dataset for semantic understanding of cloud and
  cloud shadow in sentinel-2.
\newblock {\em Scientific Data}, 9(1):782, Dec 2022.

\bibitem{beaumont2022clipretrieval}
Romain Beaumont.
\newblock Clip retrieval: Easily compute clip embeddings and build a clip
  retrieval system with them.
\newblock \url{https://github.com/rom1504/clip-retrieval}, 2022.

\bibitem{cornebise2022open}
Julien Cornebise, Ivan Orsolic, and Freddie Kalaitzis.
\newblock Open high-resolution satellite imagery: The worldstrat dataset
  {\textendash} with application to super-resolution.
\newblock In {\em Thirty-sixth Conference on Neural Information Processing
  Systems Datasets and Benchmarks Track}, 2022.

\bibitem{gal2023an}
Rinon Gal, Yuval Alaluf, Yuval Atzmon, Or Patashnik, Amit~Haim Bermano, Gal
  Chechik, and Daniel Cohen-or.
\newblock An image is worth one word: Personalizing text-to-image generation
  using textual inversion.
\newblock In {\em The Eleventh International Conference on Learning
  Representations}, 2023.

\bibitem{hu2022lora}
Edward~J Hu, yelong shen, Phillip Wallis, Zeyuan Allen-Zhu, Yuanzhi Li, Shean
  Wang, Lu Wang, and Weizhu Chen.
\newblock Lo{RA}: Low-rank adaptation of large language models.
\newblock In {\em International Conference on Learning Representations}, 2022.

\bibitem{Johnson2021}
J. Johnson, M. Douze, and H. Jegou.
\newblock Billion-scale similarity search with gpus.
\newblock {\em IEEE Transactions on Big Data}, 7(03):535--547, jul 2021.

\bibitem{maini2023tmars}
Pratyush Maini, Sachin Goyal, Zachary~C. Lipton, J.~Zico Kolter, and Aditi
  Raghunathan.
\newblock T-mars: Improving visual representations by circumventing text
  feature learning, 2023.

\bibitem{languageModel}
Luca Papariello.
\newblock xlm-roberta-base-language-detection.
\newblock
  \url{https://huggingface.co/papluca/xlm-roberta-base-language-detection},
  2021.

\bibitem{Radenovic_2023_CVPR}
Filip Radenovic, Abhimanyu Dubey, Abhishek Kadian, Todor Mihaylov, Simon
  Vandenhende, Yash Patel, Yi Wen, Vignesh Ramanathan, and Dhruv Mahajan.
\newblock Filtering, distillation, and hard negatives for vision-language
  pre-training.
\newblock In {\em Proceedings of the IEEE/CVF Conference on Computer Vision and
  Pattern Recognition (CVPR)}, pages 6967--6977, June 2023.

\bibitem{Radford2021learning}
Alec Radford, Jong~Wook Kim, Chris Hallacy, Aditya Ramesh, Gabriel Goh,
  Sandhini Agarwal, Girish Sastry, Amanda Askell, Pamela Mishkin, Jack Clark,
  et~al.
\newblock Learning transferable visual models from natural language
  supervision.
\newblock In {\em International conference on machine learning}, pages
  8748--8763. PMLR, 2021.

\bibitem{Ramesh2021}
Aditya Ramesh, Mikhail Pavlov, Gabriel Goh, Scott Gray, Chelsea Voss, Alec
  Radford, Mark Chen, and Ilya Sutskever.
\newblock Zero-shot text-to-image generation.
\newblock In Marina Meila and Tong Zhang, editors, {\em Proceedings of the 38th
  International Conference on Machine Learning}, volume 139 of {\em Proceedings
  of Machine Learning Research}, pages 8821--8831. PMLR, 18--24 Jul 2021.

\bibitem{Rombach_2022_CVPR}
Robin Rombach, Andreas Blattmann, Dominik Lorenz, Patrick Esser, and Bj\"orn
  Ommer.
\newblock High-resolution image synthesis with latent diffusion models.
\newblock In {\em Proceedings of the IEEE/CVF Conference on Computer Vision and
  Pattern Recognition (CVPR)}, pages 10684--10695, June 2022.

\bibitem{Ruiz_2023_CVPR}
Nataniel Ruiz, Yuanzhen Li, Varun Jampani, Yael Pritch, Michael Rubinstein, and
  Kfir Aberman.
\newblock Dreambooth: Fine tuning text-to-image diffusion models for
  subject-driven generation.
\newblock In {\em Proceedings of the IEEE/CVF Conference on Computer Vision and
  Pattern Recognition (CVPR)}, pages 22500--22510, June 2023.

\bibitem{Saharia2022}
Chitwan Saharia, William Chan, Saurabh Saxena, Lala Li, Jay Whang, Emily~L
  Denton, Kamyar Ghasemipour, Raphael Gontijo~Lopes, Burcu Karagol~Ayan, Tim
  Salimans, Jonathan Ho, David~J Fleet, and Mohammad Norouzi.
\newblock Photorealistic text-to-image diffusion models with deep language
  understanding.
\newblock In S. Koyejo, S. Mohamed, A. Agarwal, D. Belgrave, K. Cho, and A. Oh,
  editors, {\em Advances in Neural Information Processing Systems}, volume~35,
  pages 36479--36494. Curran Associates, Inc., 2022.

\bibitem{Schuhmann2022laionb}
Christoph Schuhmann, Romain Beaumont, Richard Vencu, Cade~W Gordon, Ross
  Wightman, Mehdi Cherti, Theo Coombes, Aarush Katta, Clayton Mullis, Mitchell
  Wortsman, Patrick Schramowski, Srivatsa~R Kundurthy, Katherine Crowson,
  Ludwig Schmidt, Robert Kaczmarczyk, and Jenia Jitsev.
\newblock {LAION}-5b: An open large-scale dataset for training next generation
  image-text models.
\newblock In {\em Thirty-sixth Conference on Neural Information Processing
  Systems Datasets and Benchmarks Track}, 2022.

\bibitem{Sumbul2019}
Gencer Sumbul, Marcela Charfuelan, Begüm Demir, and Volker Markl.
\newblock Bigearthnet: A large-scale benchmark archive for remote sensing image
  understanding.
\newblock In {\em IGARSS 2019 - 2019 IEEE International Geoscience and Remote
  Sensing Symposium}, pages 5901--5904, 2019.

\end{thebibliography}

}

\appendix
\section{LAION-EO: Version 0}

    The prototype version (version 0) of LAION-EO is accessible in the form of a .csv containing the metadata of the included samples. The anchor samples were obtained from the CloudSEN12 training subset of the cloud-free samples with high-quality or scribble annotations. No augmentations have been used, and hence, the total number of anchor samples was equal to 3,456 and the 100 nearest neighbours were queried (28,292 unique matches). When filtered for high image similarity (threshold of 0.78026) and text similarity (threshold of 0.14919), 24,933 samples are available. The metadata is available at \url{https://huggingface.co/datasets/mikonvergence/LAION-EO/tree/main/ver-0}.

\section{LAION-EO: Version 1}

    The version 1 of LAION-EO is accessible in the form of a .csv containing the metadata of the included samples. The anchor samples were obtained from the CloudSEN12 training subset of the cloud-free samples with high-quality or scribble annotations. No augmentations have been used, and hence, the total number of anchor samples was equal to 3,456 and the 1,000 nearest neighbours were queried (182,476 unique matches). When filtered for high image similarity (threshold of 0.78026) and text similarity (threshold of 0.14919), 112,985 samples are available. The metadata is available at \url{https://huggingface.co/datasets/mikonvergence/LAION-EO/tree/main/ver-1}.

\end{document}